\documentclass[conference]{IEEEtran}

\usepackage{times}
\usepackage{cite}
\usepackage{graphicx}
\usepackage{amsmath,amssymb}
\usepackage{tabularx}
\newcolumntype{Y}{>{\centering\arraybackslash}X} % Custom centered X column

\usepackage[ruled,vlined]{algorithm2e}
\DontPrintSemicolon
\SetKwInput{KwIn}{Input}
\SetKwInput{KwOut}{Output}

\usepackage{tikz}
\usetikzlibrary{arrows.meta,positioning,patterns}

\title{\LARGE MIND-CAVs: Multi-Intelligence Negotiation and Decision System for CAVs based on Intent-Driven Autonomy
}

\author{
\IEEEauthorblockN{
Mainak Mondal,
Yihang Feng,
Yangchao Luo,
Song Han
}

\IEEEauthorblockA{
University of Connecticut\\
Email(s): \{mainak.mondal, yihang.feng, yangchao.luo, song.han\}@uconn.edu
}
}

\newif\ifshowcomments
\showcommentstrue

\ifshowcomments
  \newcommand{\song}[1]{{\bf \color{red}{(SH: #1)}}}
  \newcommand{\mainak}[1]{{\bf \color{blue}{(MM: #1)}}}
\else
  \newcommand{\song}[1]{}
  \newcommand{\mainak}[1]{}
\fi

\newcommand\eat[1]{}

\begin{document}

\maketitle

\begin{abstract}
Modern autonomous vehicles largely operate as isolated agents: they rely on on-board perception and decision modules and broadcast Basic Safety Messages (BSMs) that expose only low-level kinematic state. While existing cooperative driving frameworks enable limited sensor sharing, they rarely communicate high-level maneuver intentions, and edge computing is primarily used for content delivery rather than decision arbitration. As a result, current connected autonomy lacks a principled mechanism for making globally consistent, intent-aware coordination decisions across vehicles.
To address this gap, we propose MIND-CAVs, a Multi-Intelligence Negotiation and Decision framework for connected autonomous vehicles (CAVs) based on intent-driven autonomy. Each vehicle abstracts raw sensor observations into structured intent representations, exchanges them over V2X links, and receives globally consistent coordination plans from roadside edge servers. Edge agents combine learned and rule-based arbitration mechanisms to negotiate conflicting intents among vehicles, while a cloud platform records decisions for auditing and continual retraining.
We implement MIND-CAVs in a CARLA-based AI-in-the-loop platform and evaluate it in multi-lane highway scenarios involving conflicting maneuvers and route-constrained exits. Experimental results show improved maneuver completion time and reduced unsafe proximity and unnecessary braking compared with isolated autonomy, first-come-first-served arbitration, and multi-agent reinforcement learning baselines.

% We implement an AI-in-the-loop simulation platform based on CARLA and evaluate MIND-CAVs under complex multi-vehicle traffic scenarios. Experimental results show that MIND-CAVs significantly reduces collision rates and improves traffic throughput compared to isolated autonomy, demonstrating the effectiveness of intent negotiation and multi-layer intelligence for cooperative autonomous driving.

%Autonomous vehicles primarily operate as isolated agents, broadcasting kinematic state through standard V2X messages without exposing high-level maneuver intent. As a result, vehicles must infer future behavior from instantaneous motion, leading to conservative or unstable coordination in multi-agent settings.

%We present MIND-CAVs, a Multi-Intelligence Negotiation and Decision framework that enables structured intent exchange and real-time edge arbitration for connected autonomous vehicles. Vehicles generate interpretable maneuver intents and multi-stage plans, transmit them to a roadside Multi-Access Edge Computing (MEC) node, and execute only safety-gated approvals. The architecture integrates onboard reasoning, low-latency conflict resolution, and structured audit logging for accountability and deterministic replay.

\end{abstract}

\IEEEpeerreviewmaketitle

\section{Introduction}
\label{sec:introduction}

Autonomous vehicles (AVs) today operate as highly capable but largely isolated cyber-physical systems. Although modern vehicles broadcast kinematic state through standardized Vehicle-to-Everything (V2X) messages such as Basic Safety Messages (BSMs) and Cooperative Awareness Messages (CAMs), these protocols expose only instantaneous position and motion information and do not convey future maneuver intent~\cite{monteuuis2022v2x}. Consequently, each vehicle must infer the behavior of surrounding agents from short-horizon observations and independently select actions. In dense multi-vehicle environments, this reactive paradigm often results in conservative driving, oscillatory behaviors, unnecessary braking, and fragile safety margins.

A substantial body of research has sought to improve coordination among connected autonomous vehicles (CAVs). For example, cooperative perception frameworks extend sensing range by sharing raw or feature-level sensor data~\cite{huang2023cooperative}, while centralized schedulers and cloud-assisted planners coordinate vehicle trajectories through infrastructure nodes~\cite{gao2025v2x}. Surveys have also summarized collaborative driving architectures spanning platooning, intersection management, and distributed planning~\cite{malik2021collaborative}. However, many of these approaches either (i) rely on centralized planners that are difficult to integrate with deployed V2X stacks, (ii) exchange intent information without structured negotiation or arbitration mechanisms, or (iii) focus primarily on perception and communication efficiency rather than on multi-vehicle decision governance.

At the same time, Multi-Access Edge Computing (MEC) has emerged as a key enabler of low-latency vehicular services~\cite{abdullah2023edge,5gaa2017edge}. However, in most existing architectures, MEC primarily supports content distribution, map services, or perception sharing rather than explicit multi-agent maneuver arbitration. Moreover, even when coordination mechanisms are proposed, they rarely provide an auditable record explaining why a maneuver was approved, what alternatives were rejected, and how conflicts were resolved.

This lack of structured accountability presents a critical deployment challenge. Current autonomy stacks log sensor data and control outputs, but they do not record high-level intent representations, arbitration rationales, or approval chains. As a result, autonomous driving systems remain opaque, complicating regulatory validation, incident investigation, and public trust. A scalable coordination framework must therefore satisfy not only safety and latency requirements, but also transparency and reproducibility.

To address these gaps, we propose \textbf{MIND-CAVs}, a \textit{Multi-Intelligence Negotiation and Decision System} for CAVs based on intent-driven autonomy. MIND-CAVs introduces a hierarchical Vehicle-MEC-Cloud architecture in which vehicles generate structured maneuver intents and multi-stage plans, edge (MEC) agents arbitrate and negotiate conflicting intents under bounded latency, and cloud services analyze structured logs to support learning and auditability. Unlike existing cooperative driving systems, MIND-CAVs treats coordination, real-time arbitration, and accountability as architectural primitives within a deployable V2X framework. The key contributions of this work are summarized below:

\noindent \textbf{\textbullet \;Intent-Driven Negotiation Architecture:}
We introduce a hierarchical Vehicle–MEC–Cloud coordination architecture that enables explicit intent exchange, conflict detection, priority-based arbitration, and sub-second negotiation for multi-vehicle maneuvers.

\noindent \textbf{\textbullet \;Accountability-by-Design Autonomy:}
We design a structured decision logging and audit pipeline that records intents, approvals, rejections, rationales, and outcomes, enabling deterministic replay and regulatory inspection.

\noindent \textbf{\textbullet \;AI-in-the-Loop Evaluation Platform:}
We implement MIND-CAVs in a CARLA-based AI-in-the-loop platform with live telemetry, real-time intent generation, and dataset pipelines, demonstrating improved safety and efficiency compared with isolated autonomy and learning-based baselines.

\section{Related Work}
\label{sec:related_work}

Current connected vehicle deployments primarily rely on state-based awareness messaging rather than explicit maneuver negotiation. In the United States, SAE J2735 defines Basic Safety Messages (BSMs) that broadcast position, velocity, and motion state at high frequency~\cite{sae_j2735}. Similarly, ETSI Cooperative Awareness Messages (CAMs) in Europe disseminate kinematic information to maintain vehicle awareness~\cite{etsi_cam}. While these standards improve situational awareness, they do not convey high-level maneuver intent or multi-stage plans. As a result, vehicles must infer future behavior from instantaneous motion alone, which becomes brittle in dense multi-agent interactions. This state-only paradigm motivates the \emph{Isolated Autonomy (IA)} baseline used in our evaluation, where vehicles rely solely on local perception and kinematic inference without structured intent exchange.

To move beyond state awareness, prior research has explored decentralized and cooperative maneuver coordination. Reservation-based mechanisms, such as the seminal work by Dresner and Stone~\cite{dresner2004multiagent}, allocate conflict-free space-time slots to vehicles at intersections. Subsequent studies extended cooperative maneuver concepts to decentralized coordination and infrastructure-assisted arbitration~\cite{maksimovski_survey, transaid_infra_mcs}. The ETSI Maneuver Coordination Service (MCS) further formalizes negotiation-oriented coordination at the protocol level~\cite{etsi_mcs_tr}, while extensions to maneuver coordination protocols introduce richer negotiation primitives and cost-aware decision logic~\cite{mertens_mcp_ext}. Cooperative lane-change studies also show that V2X-enabled synchronization and negotiation can reduce unnecessary braking and improve throughput~\cite{lombard_cooperative_lanechange, shah_messenger_lanechange}. However, many of these approaches rely on heuristic serialization, assume compliant participants, or lack explicit runtime arbitration with structured decision logging. These methods motivate our \emph{First-Come-First-Serve (FCFS)} baseline, representing heuristic arbitration without semantic intent reasoning or auditable decision traces.

Learning-based coordination has gained increasing attention, particularly through multi-agent reinforcement learning (MARL). Cooperative lane-change and merging problems have been formulated as MARL tasks where shared rewards encourage emergent coordination in mixed traffic~\cite{zhou_marl_lanechange_springer}. While such approaches can learn effective behaviors in simulation, they often require centralized training, carefully tuned reward functions, and assumptions about policy stationarity. Moreover, coordination remains implicit within learned policies, limiting the transparency and interpretability of runtime decisions. This class of approaches motivates the \emph{MARL} baseline in our evaluation, representing learned coordination without explicit intent negotiation or structured arbitration outcomes.

Edge and Multi-Access Edge Computing (MEC) architectures have been proposed to support low-latency vehicular applications under 3GPP NR-V2X standards~\cite{harounabadi_nr_v2x, etsi_tr_137985}. NR-V2X enhancements provide improved sidelink reliability and latency for cooperative driving~\cite{ali_nr_v2x_overview}, and recent work explores joint sensing, communication, and computation frameworks that leverage MEC for vehicular systems~\cite{li_isac_mec}. While these studies emphasize communication efficiency and computational offloading, they typically focus on perception sharing or throughput optimization rather than structured maneuver arbitration across multiple vehicles.

Finally, the need for transparency and accountability in autonomous driving has motivated research in explainable AI and interpretable decision systems~\cite{xai_survey_2024}. Efforts to generate intelligible explanations for autonomous vehicle behavior highlight the importance of structured rationales in safety-critical systems~\cite{omeiza_iv_2021}. However, such explanation mechanisms are often treated as post-hoc visualization tools rather than core coordination artifacts.

In contrast to state-only messaging, heuristic serialization, and implicit learning-based coordination, MIND-CAVs introduces structured, machine-readable intent exchange with explicit MEC-based safety gating under bounded latency constraints. By externalizing negotiation as an auditable protocol that returns ACK/PLAN/NACK decisions and logs arbitration rationales, the proposed framework unifies semantic coordination, real-time arbitration, and accountability within a deployable edge-assisted architecture.

\section{Problem Statement}
\label{sec:problem_statement}

We consider a connected autonomous driving environment comprising a finite set of vehicles 
$\mathcal{V}=\{v_1,\dots,v_N\}$ operating on a road network instrumented with roadside edge 
servers (MEC nodes) $\mathcal{E}=\{e_1,\dots,e_M\}$. Each vehicle is equipped with onboard perception 
and control modules and periodically broadcasts mandatory V2X kinematic messages (e.g., BSM/CAM). 
These messages expose instantaneous state but do not communicate future maneuver intent.

{
\vspace{0.05in}

\noindent\textbf{Local Observation Model:}
Each vehicle $v_i$ maintains a partial local observation (compact coordination snapshot)
\[
o_i^t = (x_i^t, \dot{x}_i^t, \ell_i^t, z_i^t),
\]
where $x_i^t$ and $\dot{x}_i^t$ denote pose and velocity,
$\ell_i^t$ encodes current lane index,
and $z_i^t$ denotes a compact scene representation (compressed top-down map image).

\vspace{0.05in}
\noindent\textbf{Intent Abstraction:}
We define an \emph{intent} as a structured semantic representation of a vehicle’s planned maneuver, \textit{i.e.},
\[
I_i^t = \langle g_i^t, a_i^t, \tau_i^t, C_i^t \rangle,
\]
where $g_i^t$ is the goal (\textit{e.g.}, target lane or exit), $a_i^t$ is the maneuver class 
(\textit{e.g.}, lane change, merge), $\tau_i^t$ is the execution horizon, and $C_i^t$ encodes safety constraints.  Each intent is accompanied by a plan $P_i^t$, defined as a finite sequence of high-level maneuver primitives over the horizon $\tau_i^t$. Intents are transmitted to MEC nodes over V2X links.

\vspace{0.05in}
\noindent\textbf{Accountability Constraint:}
For every arbitration outcome, the system generates a structured audit record
\[
R_i^t = \langle I_i^t, P_i^t, \hat{P}_i^t, o_i^t, d_i^t \rangle
\]
where $I_i^t$ is the proposed intent, $P_i^t$ the original plan,
$\hat{P}_i^t$ the approved or revised plan,
$o_i^t$ the telemetry snapshot at decision time,
and $d_i^t$ the arbitration rationale.
The audit trail must support deterministic replay and regulatory inspection.

\vspace{0.05in}
\noindent\textbf{Arbitration as a Constrained Decision Problem:}
Let $\mathcal{V}_j \subseteq \mathcal{V}$ denote the set of vehicles within the coverage region of MEC node $e_j$.
At time $t$, node $e_j$ receives:
(i) a set of intent-plan pairs 
\[
\mathcal{Z}_j^t = \{(I_i^t, P_i^t) \mid v_i \in \mathcal{V}_j\},
\]
(ii) the set of local observations
\[
\mathcal{O}_j^t = \{o_i^t \mid v_i \in \mathcal{V}_j\},
\]
% where each $o_i^t$ is a compact snapshot (telemetry + scene representation) rather than raw sensor streams; within $o_i^t$, $z_i^t$ can be implemented as a low-bandwidth top-down occupancy encoding or compressed image,
% and (iii) the database of previously approved active plans $\mathcal{D}_j^t$.

where $o_i^t$ is a compact telemetry–scene snapshot (telemetry + scene representation, not raw sensor streams), and $\mathcal{D}_j^t$ stores previously approved active plans.

The arbitration policy $\pi_j$ maps these inputs to an approved or revised decision:
\[
\hat{P}_i^t = 
\pi_j\!\left((I_i^t, P_i^t, o_i^t), \mathcal{Z}_j^t, \mathcal{O}_j^t, \mathcal{D}_j^t\right)
\]
where $\hat{P}_i^t$ corresponds to:
(i) approval (ACK), 
(ii) revision (PLAN), or 
(iii) rejection (NACK).

\vspace{0.05in}
We further define the instantaneous utility of $v_i$ (traffic efficiency) at time $t$ as
\[
U(v_i^t) = 
\alpha \, \dot{x}_i^t
- \beta \, B_i^t
- \gamma \, \Phi_i^t,
\]
where $\dot{x}_i^t$ denotes longitudinal speed (encouraging throughput), and $B_i^t$ is a braking indicator defined as
  \[
  B_i^t =
  \begin{cases}
  1, & \text{if } \dot{x}_i^{t-1} - \dot{x}_i^t > \delta_b 
  \text{ and } \min_{k \neq i} d_{ik}^t > d_{\text{conf}} \\
  0, & \text{otherwise}
  \end{cases}
  \]
  penalizing unnecessary braking; $\Phi_i^t$ is a safety penalty activated when $\min_{k \neq i} d_{ik}^t < d_{\text{safe}}$, capturing unsafe proximity.

% Here $\delta_b > 0$ is a deceleration threshold defining significant braking, $d_{\text{conf}}$ denotes a conflict radius below which braking is justified, $d_{ik}^t$ denotes Euclidean inter-vehicle distance and $\alpha, \beta, \gamma > 0$ weight efficiency–safety trade-offs.

Here $\delta_b$ is a deceleration threshold, $d_{\text{conf}}$ a conflict radius, 
$d_{ik}^t$ the inter-vehicle distance, and $\alpha,\beta,\gamma>0$ weight efficiency–safety trade-offs.

\vspace{0.05in}
\noindent\textbf{Optimization Objective~\footnote{
This objective formalizes arbitration goals; the implemented policy $\pi_j$ approximates it via constrained semantic reasoning under deterministic safety validation rather than explicit numerical optimization.
}:}
The arbitration objective is to improve traffic efficiency while respecting safety and latency constraints, which we formalize via the cumulative utility:
\[
\max_{\{\pi_j\}}
\mathbb{E}\!\left[
\sum_t \sum_{v_i \in \mathcal{V}} U(v_i^t)
\right]
\]
subject to:
(i) conflict-free maneuver execution,
(ii) bounded arbitration delay $\Delta_{\max}$~\footnote{$\Delta_{\max}$ is the maximum allowable end-to-end arbitration latency for a single intent request, measured from vehicle TX to RX of the MEC decision.},
and (iii) generation of complete structured audit records. The expectation is taken over traffic realizations and policy stochasticity.

% \noindent\textbf{Local Observation Model:}
% Each vehicle $v_i$ maintains a partial local observation
% \[
% o_i^t = (x_i^t, \dot{x}_i^t, \mathcal{N}_i^t, \ell_i^t)
% \]
% at time $t$, where:
% $x_i^t$ and $\dot{x}_i^t$ denote pose and velocity,
% $\mathcal{N}_i^t$ denotes locally perceived neighboring agents,
% and $\ell_i^t$ represents lane context (current lane index, route constraints, and adjacent-lane availability).
% Vehicles cannot directly observe the future goals or planned maneuvers of surrounding vehicles.

% \vspace{0.05in}
% \noindent\textbf{Edge Arbitration:}
% Each MEC node $e_j$ receives a set of intents $\mathcal{I}_j^t$ from vehicles within its coverage area 
% and applies an arbitration policy $\pi_j$ that produces an approved subset 
% $\hat{\mathcal{I}}_j^t \subseteq \mathcal{I}_j^t$ such that:
% (i) approved intents are mutually conflict-free,
% (ii) arbitration completes within a bounded delay $\Delta_{\max}$, and
% (iii) decisions are computable using MEC-scale resources.
\section{System Architecture}
\label{sec:architecture}

MIND-CAVs is a hierarchical coordination framework organized into three intelligence tiers:
(i) Vehicle Intelligence (VI),
(ii) Edge Intelligence (EI), and
(iii) Cloud Intelligence (CI).
Fig.~\ref{fig:mindcavs_arch} illustrates the system architecture.

\begin{figure*}[t]
    \centering
    \includegraphics[width=\textwidth]{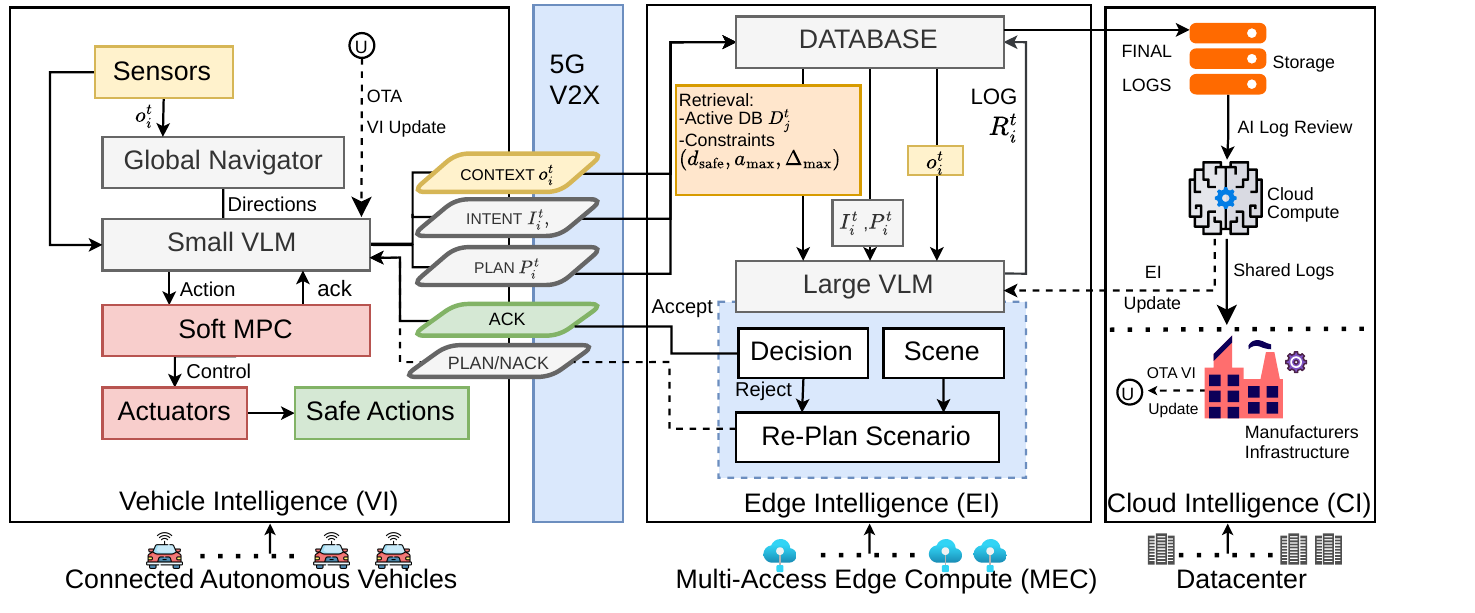}
    \vspace{-0.3in}
    \caption{MIND-CAVs architecture: vehicles generate intent–plan pairs; MEC performs arbitration and safety gating; cloud aggregates structured decision logs.}
    \label{fig:mindcavs_arch}
\end{figure*}

\vspace{-0.05in}
\subsection{VI: Intent Proposal and Execution}

Each vehicle maintains a local observation $o_i^t$. From this observation, the vehicle generates an intent-plan pair $(I_i^t, P_i^t)$.  A plan $P_i^t$ is a finite sequence of maneuver primitives (e.g., keep-lane, accelerate, initiate lane change).
The intent-plan generation mechanism is modular and may be rule-based, learned, or LLM-assisted.

Each vehicle operates under an execute-after-approval loop:
\[
(I_i^t, P_i^t, o_i^t) \rightarrow \text{MEC review} \rightarrow \hat{P}_i^t \rightarrow \text{execution}.
\]

A plan step corresponds to execution of a single maneuver primitive.
If execution exceeds its declared horizon $\tau_i^t$ or arbitration delay exceeds $\Delta_{\max}$,
the vehicle reverts to a conservative fallback policy (e.g., keep lane, reduce speed, increase headway) until a valid response is received. Additionally, we apply a soft MPC-based safety filter that enforces control-barrier-style constraints (e.g., minimum gap and bounded acceleration), preventing execution of unsafe maneuvers even when upstream intent proposals are imperfect.

\vspace{-0.05in}
\subsection{EI: Negotiation and Arbitration}

MEC nodes receive intent–plan pairs from vehicles within their coverage region and apply the arbitration policy $\pi_j$ defined in Sec.~\ref{sec:problem_statement}. 

The Edge Intelligence (EI) module performs semantic intent arbitration using a constrained VLM over the active traffic context, incorporating priority rules and plan revision when necessary. A deterministic spatial–temporal validator subsequently enforces safety constraints and conflict-free execution.

Based on the validated outcome, the MEC node acts as a safety gate and returns one of three responses: ACK (approve), PLAN (revise), or NACK (reject). All decisions are computed within bounded latency $\Delta_{\max}$ to enable sub-second coordination.

% MEC nodes receive intent-plan pairs from vehicles within their coverage region and apply the arbitration policy $\pi_j$ defined in Sec.~\ref{sec:problem_statement}.

% The Edge Intelligence (EI) module performs:
% (i) spatial-temporal conflict detection,
% (ii) priority resolution (giving precedence to emergency vehicles and otherwise following a first-come-first-served policy), and
% (iii) plan revision when required (e.g., inserting delays or adjusting speed bounds).

% Based on the arbitration outcome, the MEC node acts as a safety gate and returns one of three responses: ACK (approve), PLAN (revise), or NACK (reject). All decisions are computed within a bounded latency $\Delta_{\max}$ to enable sub-second coordination.

\vspace{-0.05in}
\subsection{CI: Audit and Fleet-Level Learning}

The cloud tier aggregates structured audit records
$R_i^t = \langle I_i^t, P_i^t, \hat{P}_i^t, o_i^t, d_i^t \rangle$. Audit records are generated at MEC nodes, stored locally during operation, and asynchronously transmitted to cloud storage as immutable structured logs.

The cloud tier does not participate in real-time control.
It supports:
(i) forensic replay,
(ii) large-scale performance analysis,
and (iii) dataset generation for improving intent and arbitration policies.

\vspace{-0.05in}
\subsection{Communication Flow}

MIND-CAVs operates over three communication paths:

\vspace{0.05in}
\noindent \textbf{Baseline V2X broadcast:}
Vehicles continue to transmit kinematic state via BSM/CAM messages.

\noindent \textbf{Intent overlay channel:}
Vehicles transmit intent–plan pairs to MEC; MEC returns approved or revised plans.

\noindent \textbf{Audit propagation channel:}
MEC nodes asynchronously transmit structured audit records to the cloud tier for persistent storage and offline analysis.

\section{MIND-CAVs Coordination Protocol and Arbitration Algorithm}
\label{sec:protocol}

We now describe the runtime coordination protocol executed between VI and EI tiers. Formal definitions of intents, plans, and audit records are given in Sec.~\ref{sec:problem_statement}. Here we focus on (i) vehicle-side intent generation, (ii) MEC arbitration, and (iii) bounded-latency negotiation.

\vspace{-0.05in}
\subsection{Vehicle-Side Intent Generation}
\label{sec:vehicle_intent}

At time $t$, vehicle $v_i$ constructs its local observation 
$o_i^t$ as defined in Sec.~\ref{sec:problem_statement}, which includes 
kinematic state, lane context, and a compact top-down scene representation $z_i^t$. 
Based on this observation and a global routing goal, the vehicle generates an intent-plan pair $(I_i^t, P_i^t)$.

The plan $P_i^t$ is a finite sequence of high-level maneuver primitives:
\[
P_i^t = \{u_i^1, u_i^2, \dots, u_i^K\},
\]
where each primitive $u_i^k$ encodes a target lane index, longitudinal speed, and hold duration over a short horizon.

In our implementation, intent-plan generation is performed using a VLM-assisted module to enable semantic goal interpretation and structured multi-stage maneuver synthesis without brittle rule-based planners. A structured prompt containing $o_i^t$ and a user-defined goal (e.g., ``move to the rightmost lane'') is used to generate $(I_i^t, P_i^t)$. Vehicle $v_i$ transmits $(I_i^t, P_i^t, o_i^t)$ to the nearest MEC node, and continues executing its previously approved plan until an arbitration response is received.

% In our implementation, intent-plan generation is performed using a VLM-assisted module. 
% We construct a structured prompt containing $o_i^t$ and a user-defined goal (e.g., ``move to the rightmost lane''), and request generation of $(I_i^t, P_i^t)$.

\vspace{-0.05in}
\subsection{MEC Arbitration Mechanism}
\label{sec:mec_arbitration}

Each MEC node $e_j$ maintains a database $\mathcal{D}_j^t$ of active approved plans for vehicles within its coverage region $\mathcal{V}_j$.
Upon receiving a request $(I_i^t, P_i^t, o_i^t)$, the MEC executes a constrained VLM-based arbitration agent. 
The agent receives as input a structured prompt containing:
(i) the incoming intent-plan pair $(I_i^t, P_i^t)$,
(ii) the compact observation snapshot $o_i^t$,
(iii) the set of currently active intents and plans in $\mathcal{D}_j^t$, and
(iv) the lexicographic priority rules.
% The agent generates one of three decisions: $\texttt{ACK}$, $\texttt{PLAN}$ (revised plan), or $\texttt{NACK}$. (shown in Algorithm \ref{alg:mec_arbitration}).
The agent outputs $\texttt{ACK}$, $\texttt{PLAN}$ (revised plan), or $\texttt{NACK}$ (Algorithm~\ref{alg:mec_arbitration}).

\subsubsection{Plan Representation}

Each plan is represented as a time-indexed occupancy function
\[
\Omega_i(t') = (\ell_i(t'), s_i(t'))
\]
where $\ell_i(t')$ is predicted lane index and $s_i(t')$ longitudinal position over $t' \in [t, t+\tau_i]$ derived from $P_i^t$.

\subsubsection{Conflict Detection}

% The MEC performs VLM-based arbitration over the full traffic context. 
% A deterministic spatial-temporal validator is applied post-hoc to ensure that any approved or revised plan satisfies minimum distance and kinematic safety constraints. 
% A conflict is detected if there exists $t'$ such that:
% \[
% \ell_i(t') = \ell_k(t') 
% \quad \text{and} \quad 
% |s_i(t') - s_k(t')| < d_{\text{safe}}
% \]
% for some vehicle $v_k \in \mathcal{V}_j$, $k \neq i$. If the VLM proposes approval (\texttt{ACK}) or a revised plan (\texttt{PLAN}), the deterministic validator verifies that the resulting plan is conflict-free; invalid outputs are rejected (\texttt{NACK}).

Arbitration is performed by the VLM over the full traffic context, followed by a deterministic spatial–temporal validator enforcing safety constraints. 
A conflict exists if there is $t'$ such that
\[
\ell_i(t') = \ell_k(t') \ \text{and}\ |s_i(t') - s_k(t')| < d_{\text{safe}}
\]
for some $v_k \in \mathcal{V}_j$, $k \neq i$. 
If the validator fails, the decision is overridden to $\texttt{NACK}$.

\begin{algorithm}[h]
\small
\caption{\small VLM-assisted Arbitration w/ Safety Guard}

\label{alg:mec_arbitration}
\SetKwInOut{Input}{Input}
\SetKwInOut{Output}{Output}
\SetKw{KwRet}{return}

\Input{Incoming request $(I_i^t, P_i^t, o_i^t)$; active database $\mathcal{D}_j^t$; constraints $(d_{\text{safe}}, a_{\max}, \Delta_{\max})$}
\Output{Decision $\in \{\texttt{ACK}, \texttt{PLAN}, \texttt{NACK}\}$ and $\hat{P}_i^t$}

\BlankLine
\textbf{/* Retrieve active context */}\;
$\mathcal{Z}_j^t \leftarrow$ active intent-plan pairs from $\mathcal{D}_j^t$\;
$\mathcal{O}_j^t \leftarrow$ observation snapshots associated with $\mathcal{Z}_j^t$\;

\BlankLine
\textbf{/* Construct arbitration prompt */}\;
$\mathsf{prompt} \leftarrow \mathsf{FormatPrompt}(I_i^t, P_i^t, o_i^t,\ \mathcal{Z}_j^t,\ \mathcal{O}_j^t,\ \text{constraints})$\;

\BlankLine
\textbf{/* Single-pass VLM arbitration */}\;
$(d_i^t, \texttt{decision}, \tilde{P}_i^t) \leftarrow \mathsf{VLM}(\mathsf{prompt})$\;

\If{$\texttt{decision} = \texttt{ACK}$}{
    $\hat{P}_i^t \leftarrow P_i^t$\;
}
\ElseIf{$\texttt{decision} = \texttt{PLAN}$}{
    $\hat{P}_i^t \leftarrow \tilde{P}_i^t$\;
}
\Else{
    $\texttt{decision} \leftarrow \texttt{NACK}$\;
    $\hat{P}_i^t \leftarrow \emptyset$\;
}

\BlankLine
\textbf{/* Deterministic safety verification */}\;
\If{$\texttt{decision} \in \{\texttt{ACK}, \texttt{PLAN}\}$}{
    \If{$\neg\,\mathsf{ValidatePlan}(\hat{P}_i^t,\ \mathcal{Z}_j^t,\ d_{\text{safe}},\ a_{\max})$}{
        $\texttt{decision} \leftarrow \texttt{NACK}$\;
        $\hat{P}_i^t \leftarrow \emptyset$\;
        $d_i^t \leftarrow d_i^t \,\|\, \text{``; rejected by safety validator''}$\;
    }
}

\BlankLine
\textbf{/* Commit + accountability logging */}\;
\If{$\texttt{decision}\in\{\texttt{ACK},\texttt{PLAN}\}$}{
    Update $\mathcal{D}_j^t$ with $(I_i^t,\hat{P}_i^t,o_i^t)$ as active\;
}
\Else{
    Record rejected request metadata in $\mathcal{D}_j^t$\;
}

Write audit record 
$R_i^t=\langle I_i^t,P_i^t,\hat{P}_i^t,o_i^t,d_i^t\rangle$\;

\KwRet $(\texttt{decision},\hat{P}_i^t)$\; 
\end{algorithm}

\subsubsection{Priority Rule}

If conflicts exist, a lexicographic priority rule is applied where vehicles with emergency class take precedence. Otherwise, earlier submission timestamps take precedence and ties are broken by vehicle identifier.  This priority rule is included as a hard constraint in the MEC prompt to ensure consistent arbitration decisions across conflicts.

\subsubsection{Plan Revision}

If $v_i$ is prioritized, a revised plan $\hat{P}_i^t$ is generated. The revision operates under a safety envelope: (i) enforce minimum longitudinal gap $d_{\text{safe}}$, (ii) enforce acceleration and deceleration bounds, and (iii) preserve goal $g_i^t$ whenever feasible.

A constrained VLM-based agent generates a single revised plan within this envelope (e.g., delay lane change, reduce speed, insert yielding phase). Only one revision attempt is permitted to preserve bounded latency. If $v_i$ is not prioritized, the proposal is rejected.

\subsubsection{Bounded Latency}

Bounded arbitration delay $\Delta_{\max}$ is achieved through (i) deterministic single-pass conflict detection, (ii) a fixed lexicographic priority rule (no iterative negotiation), (iii) at most one constrained plan revision, and (iv) bounded token generation at the MEC tier.

No multi-round negotiation loops are permitted, ensuring predictable sub-second execution.

\vspace{-0.05in}
\subsection{Timing and Latency Analysis}
\label{sec:timing}

We estimate the end-to-end negotiation latency across the Vehicle $\rightarrow$ MEC $\rightarrow$ Vehicle loop based on measured inference throughput and reported 5G V2X latency models~\cite{collperales2023v2x}. On Jetson AGX-class hardware, generating a structured intent and short plan ($\sim$20 tokens) requires approximately 270~ms using 4-bit quantization. The resulting $\sim$200~kB payload can be transmitted over MEC-assisted 5G V2X in less than 10~ms. At the MEC server, running on A100-class hardware with a measured throughput of approximately 22 tokens/s, arbitration latency is about 45~ms for an $\texttt{ACK}$ response (1 token) and $\sim$400~ms when generating a revised $\texttt{PLAN}$ ($\sim$10 tokens). The return transmission of the $\sim$10~kB response requires less than 10~ms. Combining these components yields an estimated end-to-end negotiation latency of approximately 335~ms for the $\texttt{ACK}$ case ($270 + 10 + 45 + 10$) and 690~ms for the $\texttt{PLAN}$ revision case ($270 + 10 + 400 + 10$). Even in the revision case, the total negotiation cycle remains below 700~ms. At highway speeds (100~km/h), this corresponds to roughly 20~m of travel, which remains within the short-horizon planning range typically considered by autonomous driving systems.

\eat{We estimate end-to-end negotiation latency across the Vehicle $\rightarrow$ MEC $\rightarrow$ Vehicle loop based on measured inference throughput and reported 5G V2X latency models~\cite{collperales2023v2x}.

\subsubsection{Vehicle-Side Generation}

On Jetson AGX-class hardware, structured intent and short plan generation ($\sim$20 tokens) requires approximately 270~ms under 4-bit quantization. The transmission (TX) of a $\sim$200~kB payload over MEC-based 5G V2X is modeled at less than 10~ms~\cite{collperales2023v2x}.

\subsubsection{MEC Arbitration}

On A100-class hardware, measured token throughput is approximately 22 tokens/s:

\begin{itemize}
    \item $\texttt{ACK}$ generation (1 token): $\sim$45~ms,
    \item $\texttt{PLAN}$ revision ($\sim$10 tokens): $\sim$400~ms.
\end{itemize}

Return TX latency ($\sim$10~kB) is less than 10~ms~\cite{collperales2023v2x}.

\subsubsection{End-to-End Delay}

\begin{itemize}
    \item \textbf{ACK case : } $270 + 10 + 45 + 10 \approx 335~\text{ms}$
    
    \item \textbf{PLAN case : } $270 + 10 + 400 + 10 \approx 690~\text{ms}$

\end{itemize}

Even in the revision case, the total negotiation cycle remains below 700~ms. At highway speeds (100~km/h), this corresponds to approximately 20~m of travel, which remains within typical short-horizon planning ranges in autonomous driving systems.}
\section{Experimental Setup and Evaluation Methodology}
\label{sec:evaluation}

We evaluate MIND-CAVs in a controlled highway driving environment designed to isolate coordination behavior and intent negotiation under minimal confounding factors. %All experiments are conducted in a straight, four-lane highway segment with no intersections and consistent lane geometry. 
Each scenario is executed 10 times with identical initial configurations but independent stochastic VLM sampling.

\subsection{Simulation Environment and Platform}
\label{sec:simulation_environment}
% (AutoLLM \cite{autollm_github})
All experiments are conducted using CARLA, integrated with our custom \textit{MIND-CAVs Simulation Conductor}, which provides an AI-in-the-loop orchestration layer for multi-vehicle coordination, monitoring, and intent-based control (see Fig.~\ref{fig:mindcavs_sim}).
The MIND-CAV simulator  manages CARLA synchronization (Fig.~\ref{fig:mindcavs_sim}a), vehicle spawning (Fig.~\ref{fig:mindcavs_sim}b), intent exchange (Fig.~\ref{fig:mindcavs_sim}c), MEC arbitration, and structured logging (Fig.~\ref{fig:mindcavs_sim}d). The experiments use synchronous stepping and persistent JSON logs for telemetry, intent records, and arbitration outcomes. Simulation seeds, initial vehicle states, and scenario configurations are recorded to ensure deterministic replay and reproducibility. 
The simulation platform runs on a workstation equipped with an Intel i7-10700 CPU (2.90~GHz), 64~GB RAM, and an NVIDIA RTX~3070 GPU. The system runs Ubuntu~24.04 LTS and CARLA~0.9.15.

% \subsubsection{Hardware and Software Configuration}
% The simulation platform runs on a workstation equipped with an Intel i7-10700 CPU (2.90~GHz), 64~GB RAM, and an NVIDIA RTX~3070 GPU. The system runs Ubuntu~24.04 LTS and CARLA~0.9.15. This configuration supports real-time simulation with synchronous stepping, multi-camera streaming, and concurrent VLM-based intent generation without dropped frames or control lag.

\subsubsection{CARLA Configuration}
CARLA is operated in synchronous mode with a fixed simulation timestep to ensure deterministic execution across runs. Vehicles are spawned programmatically using predefined spawn points on a straight four-lane highway in Town\_04 (map provided by CARLA). All vehicles use identical dynamics models, control limits, and sensor configurations. No background traffic is introduced, ensuring that observed behaviors arise solely from interactions between coordinated vehicles.

\begin{figure*}[t]
    \centering
    \vspace{-0.1in}
    \includegraphics[width=\textwidth]{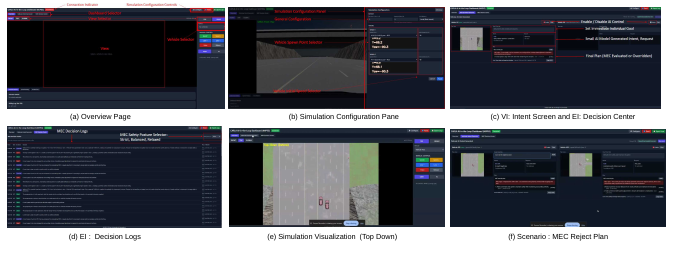}
    \vspace{-0.4in}
    \caption{AI-in-the-loop simulation platform used for evaluation}
    \vspace{-0.15in}
    \label{fig:mindcavs_sim}
\end{figure*}

\subsubsection{Intent Execution and Control Flow}
Vehicles generate intents and multi-stage plans using an onboard VLM, which consumes top-down imagery, vehicle telemetry, and lane context. Intents are transmitted to the MEC arbitration module via the backend. Until an acknowledgement (ACK) or revised plan (PLAN) is received, vehicles continue executing their previously approved plan or a default conservative policy (keep-lane with speed maintenance).

% All MEC decisions, including approvals, rejections, and plan revisions, are logged alongside the corresponding intents and telemetry snapshots. This enables deterministic replay of every experiment and supports auditability and reproducibility.

\subsection{Baseline Methods}

To rigorously evaluate the impact of intent-driven arbitration and accountability, we compare MIND-CAVs against three representative coordination paradigms from prior literature.

\vspace{0.025in}
\noindent {\bf Isolated Autonomy (IA).}  
Vehicles operate independently without intent exchange or MEC arbitration. Each vehicle relies solely on onboard perception and kinematic inference from neighboring agents and executes maneuvers greedily subject to local safety constraints. This baseline reflects current production autonomous driving stacks that do not perform explicit multi-vehicle negotiation.

\vspace{0.025in}
\noindent {\bf Centralized First-Come-First-Serve (FCFS) Arbitration.}  
A deterministic centralized controller resolves maneuver conflicts using a FCFS priority rule. Vehicles submit maneuver requests, and conflicts are resolved based on arrival order without negotiation or plan revision. This baseline is inspired by reservation-based intersection management and centralized scheduling approaches~\cite{dresner2004multiagent}. It represents classical conflict-free arbitration without intent reasoning or accountability logging.

\vspace{0.025in}
\noindent {\bf Multi-Agent Reinforcement Learning (MARL).}  
A multi-agent policy observes shared vehicle states and outputs maneuver decisions optimized to minimize delay and safety violations. Coordination emerges implicitly through learned value functions, as commonly studied in cooperative autonomous driving and intersection management~\cite{samak2023multiagent}. However, no explicit intent negotiation or structured audit logs are produced.

% We trained a small MLP policy by supervising it on heuristic labels. Synthetic samples were generated by randomizing ego/neighbor lanes, speeds, and distances.
% The heuristic chooses lane‑change actions when safe, otherwise holds or adjusts speed to maintain a 50 km/h flow target. The model (2‑layer MLP, 32–16 hidden units, ReLU) is trained for 200 iterations and saved as a discrete action classifier.

% All baselines operate under identical vehicle dynamics, sensor configurations, and initial conditions to ensure fair comparison with MIND-CAVs.

\subsection{Scenario Definitions}

\vspace{0.025in}
\noindent \textbf{Scenario 1 ($S1$) : Single-Agent Multi-Lane Change.}  
Two vehicles are initialized side-by-side: one in the leftmost lane (Lane~1) and one in the adjacent lane to the right (Lane~2). Both vehicles travel at 40~km/h. The vehicle in Lane~1 is tasked to move to the rightmost lane (Lane~4), while the vehicle in Lane~2 maintains its lane. This scenario evaluates whether MIND-CAVs enables safe and efficient multi-stage lane changes w/o inducing unnecessary braking or deadlock.

\vspace{0.025in}
\noindent \textbf{Scenario 2 ($S2$) : Conflicting Lane Change Intentions.}  
The initial configuration is identical to Scenario~1. However, the vehicle in Lane~1 intends to move to the rightmost lane (Lane~4), while the vehicle in Lane~2 simultaneously intends to move to the leftmost lane (Lane~1). This scenario introduces direct intent conflict and evaluates the effectiveness of MEC-based arbitration, priority resolution, and negotiation in preventing unsafe simultaneous maneuvers.

\vspace{0.025in}
\noindent \textbf{Scenario 3 ($S3$) : Route-Constrained Exit Maneuver.}  
The setup mirrors Scenario~1, but the vehicle in the leftmost lane is assigned a route-level goal requiring it to take an exit on the right within the next 60 seconds. The second vehicle maintains its lane. This scenario evaluates long-horizon intent planning and the system’s ability to sequence intermediate maneuvers (multi-lane changes) while maintaining safety and efficiency over extended time horizons.

% \subsection{Evaluation Metrics}
% Performance is evaluated using the following metrics:

% \begin{itemize}
%     \item \textbf{Average maneuver completion time:} time required for a vehicle to complete its intended maneuver or reach the designated exit.
%     \item \textbf{Safety violations:} number of collisions or minimum-gap violations observed during execution.
%     \item \textbf{Unnecessary braking events:} count of braking actions not required by safety constraints, indicating coordination inefficiency.
%     \item \textbf{Fairness:} relative delay imposed on non-target vehicles due to coordination decisions.
%     \item \textbf{Audit completeness:} percentage of maneuvers for which a complete intent–decision–outcome trace is recorded.
% \end{itemize}

% All metrics are averaged across 10 runs per scenario. For MIND-CAVs, we additionally analyze arbitration outcomes (ACK, PLAN revision, NACK) and decision rationales to assess transparency and accountability.

\subsection{Evaluation Metrics}
\label{sec:metrics}

We employ three important metrics in the performance evaluation. All metrics are computed offline from per-run telemetry logs sampled at approximately 10~Hz. 
%For each scenario and coordination mode, results are reported as mean $\pm$ standard deviation ($\mu \pm \sigma$) and 95\% confidence intervals ($CI_{95\%}$) across runs.

\vspace{0.025in}
\noindent {\bf Maneuver Completion Time} is defined as the elapsed time between the first telemetry timestamp and the moment all target vehicles satisfy their goal condition. In S1 and S2, the goal condition is that the target vehicle reaches the designated goal lane. While in S3, the goal condition is that the target vehicle reaches the exit region within the specified route horizon. If the goal is not achieved before the predefined timeout, the completion time is set equal to the timeout value.

\vspace{0.025in}
\noindent {\bf Gap Violation} occurs when the Euclidean distance between any pair of vehicles at a telemetry frame is below the safety threshold $d_{\text{safe}} = 5$~m.  Formally, at telemetry frame $t$, for vehicles $i$ and $k$, their distance ($d_{ik}^t$) is $d_{ik}^t = \| x_i^t - x_k^t \|$.  If $d_{ik}^t < d_{\text{safe}}$, a violation counter is incremented. 

Violations are accumulated across all telemetry frames in all runs for a scenario. This metric captures unsafe proximity events during maneuver execution, and corresponds to activation of the safety penalty term $\Phi_i^t$ defined in Sec.~\ref{sec:problem_statement}.

\vspace{0.025in}
\noindent {\bf Unnecessary Braking} serves as a proxy for avoidable deceleration due to coordination inefficiencies rather than safety-critical responses. Unnecessary braking events are counted when a vehicle’s speed decreases more than $\delta_b = 5$~km/h between consecutive telemetry frames, and no other vehicle is within a conflict distance $d_{\text{conf}} = 50$~m.

Formally, if: 
$v_i^{t-1} - v_i^t > \delta_b
 \text{ and } 
\min_{\substack{v_k \in \mathcal{V} \\ k \ne i}} d_{ik}^t > d_{\text{conf}}$, then an unnecessary braking event is recorded.

% \vspace{-0.2in}
% \[
% v_i^{t-1} - v_i^t > \delta_b
% \quad \text{and} \quad
% \min_{\substack{v_k \in \mathcal{V} \\ k \ne i}} d_{ik}^t > d_{\text{conf}},
% \]

\section{Evaluation Results}
\label{sec:results}

We evaluate MIND-CAVs against three baselines across the three scenarios described in Sec.~\ref{sec:evaluation}. Each configuration uses identical initial states and vehicle dynamics, with stochasticity arising from independent VLM sampling. Metrics reported are mean ($\mu$) $\pm$ standard deviation ($\sigma$) and confidence intervals ($CI_{95\%}$) across runs, summarized in table~\ref{tab:combined_table}.

% \subsection{Scenario S1: Multi-Lane Change to Rightmost Lane}

% MIND-CAVs achieves the lowest completion time (29.10 $\pm$ 1.17 s), improving over IA (38.57 $\pm$ 2.70 s) by 24.6\% and over MARL (40.87 $\pm$ 5.23 s) by 28.8\%. 

% In addition, MIND-CAVs eliminates both gap violations and unnecessary braking (0.0 $\pm$ 0.0 in both metrics), whereas all baselines exhibit non-zero safety events. These results indicate that MEC-mediated approval prevents premature or conflicting lane-change maneuvers while preserving flow efficiency.

\subsection{Scenario S1: Multi-Lane Change to Rightmost Lane}

In S1, MIND-CAVs achieves the lowest completion time (29.10 $\pm$ 1.17 s), outperforming IA (38.57 $\pm$ 2.70 s), FCFS (40.15 $\pm$ 3.30 s), and MARL (40.87 $\pm$ 5.23 s). The reduction corresponds to approximately 24-29\% improvement over the baselines.

MIND-CAVs exhibits zero observed gap violations and zero unnecessary braking events across all runs. In contrast, FCFS, IA, and MARL all incur non-zero safety events and braking instances. Notably, MARL and FCFS show higher braking frequencies (1.50 $\pm$ 0.71 and 1.10 $\pm$ 0.74 respectively), indicating less stable maneuver sequencing. These results suggest that intent-aware arbitration enables efficient multi-stage lane changes while preventing premature or conflicting maneuvers.

\begin{figure}[h!]
    \centering
    \includegraphics[height=1.3in]{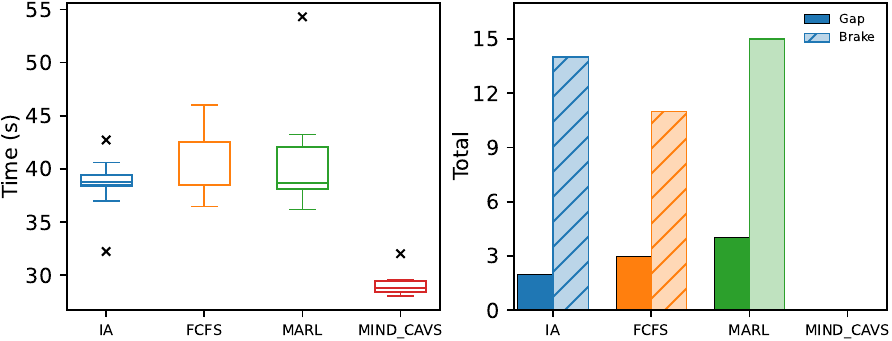}
    \vspace{-0.1in}
    \caption{Scenario S1 Results. Left: Completion time distribution. Right: Total gap‑violation and unnecessary‑brake counts.}
    \label{fig:S1_results}
    \vspace{-0.1in}
\end{figure}

% \subsection{Scenario S2: Opposing Lane Changes}

% In the conflicting lane-change scenario, MIND-CAVs has the lowest completion time (39.46 $\pm$ 4.84 s), outperforming IA (44.69 $\pm$ 5.25 s) and MARL (44.88 $\pm$ 5.46 s).

% Gap violations are comparable across methods in this scenario. However, MIND-CAVs reduces unnecessary braking relative to IA and MARL, indicating more stable conflict resolution under centralized review.

\subsection{Scenario S2: Opposing Lane Changes}

In S2, MIND-CAVs again achieves the lowest completion time (39.46 $\pm$ 4.84 s), compared to IA (44.69 $\pm$ 5.25 s), MARL (44.88 $\pm$ 5.46 s), and FCFS (46.13 $\pm$ 7.53 s).

Gap violations are similar across all methods (approximately 0.20 $\pm$ 0.63), indicating that this scenario is primarily resolved through timing rather than safety-critical conflicts. However, unnecessary braking is reduced under MIND-CAVs (0.40 $\pm$ 0.70) relative to IA (1.70 $\pm$ 1.16) and MARL (1.50 $\pm$ 1.08). FCFS shows fewer braking events than IA but exhibits higher variance in completion time. These results indicate that centralized intent arbitration improves maneuver stability under direct conflict without sacrificing safety.

\begin{figure}[h!]
    \centering
    \includegraphics[height=1.3in]{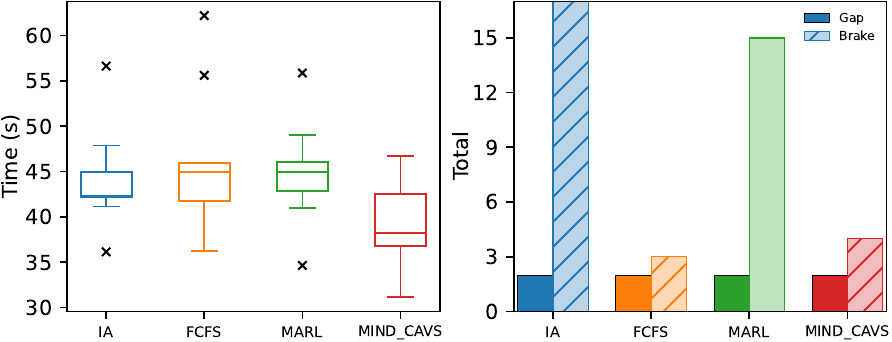}
    \caption{Scenario S2 Results. Left: Completion time distribution. Right: Total gap‑violation and unnecessary‑brake counts.}
    \label{fig:S2_results}
    \vspace{-0.1in}
\end{figure}

% \subsection{Scenario S3: Route-Constrained Exit}

% In the route-constrained exit scenario, MIND-CAVs achieves the lowest completion time (39.56 $\pm$ 2.49 s) and substantially reduces gap violations (0.20 $\pm$ 0.63) compared to IA (1.60 $\pm$ 0.84) and FCFS (1.20 $\pm$ 1.03).

% These results suggest that intent arbitration enables safe multi-stage lane transitions toward long-horizon route goals without degrading traffic flow.

\subsection{Scenario S3: Route-Constrained Exit}

In S3, MIND-CAVs achieves the lowest completion time (39.56 $\pm$ 2.49 s), compared to IA (40.73 $\pm$ 4.71 s), FCFS (42.04 $\pm$ 5.67 s), and MARL (41.32 $\pm$ 8.64 s).

Gap violations are reduced under MIND-CAVs (0.20 $\pm$ 0.63) relative to IA (1.60 $\pm$ 0.84), FCFS (1.20 $\pm$ 1.03), and MARL (1.00 $\pm$ 1.05). Unnecessary braking is also lower under MIND-CAVs (0.60 $\pm$ 0.84) compared to IA (1.60 $\pm$ 2.07), though MARL shows moderate braking levels (0.90 $\pm$ 1.37). These findings suggest that intent arbitration supports safe multi-stage lane sequencing toward long-horizon goals while maintaining comparable or improved efficiency.

\begin{figure}[h!]
    \centering
    \includegraphics[height=1.3in]{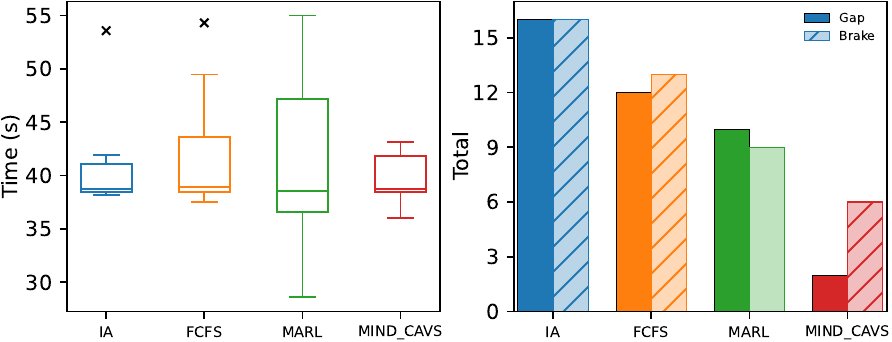}
    \caption{Scenario S3 Results. Left: Completion time distribution. Right: Total gap‑violation and unnecessary‑brake counts.}
    \label{fig:S3_results}
    \vspace{-0.1in}
\end{figure}

\begin{table*}[h!]
\centering
\caption{Statistical analysis of Completion Time, Gap Violations and Unnecessary Braking across scenarios}
\scriptsize
\begin{tabularx}{0.8\textwidth}{|l||Y|Y||Y|Y||Y|Y|}
\hline
\multicolumn{7}{|c|}{\bf{COMPLETION TIME}} \\
\hline
 & \multicolumn{2}{c||}{S1} & \multicolumn{2}{c||}{S2} & \multicolumn{2}{c|}{S3} \\
\cline{2-7}
 & $\mu\pm\sigma$ & $CI_{95\%}$ & $\mu\pm\sigma$ & $CI_{95\%}$ & $\mu\pm\sigma$ & $CI_{95\%}$ \\
\hline
\textbf{FCFS} & $40.15 \pm 3.30$ & $[37.79, 42.50]$ & $46.13 \pm 7.53$ & $[40.74, 51.51]$ & $42.04 \pm 5.67$ & $[37.98, 46.09]$ \\
\hline
\textbf{IA} & $38.57 \pm 2.70$ & $[36.64, 40.50]$ & $44.69 \pm 5.25$ & $[40.93, 48.44]$ & $40.73 \pm 4.71$ & $[37.36, 44.10]$ \\
\hline
\textbf{MARL} & $40.87 \pm 5.23$ & $[37.12, 44.61]$ & $44.88 \pm 5.46$ & $[40.98, 48.79]$ & $41.32 \pm 8.64$ & $[35.14, 47.50]$ \\
\hline
\textbf{MIND-CAVs} & $29.10 \pm 1.17$ & $[28.26, 29.94]$ & $39.46 \pm 4.84$ & $[36.00, 42.93]$ & $39.56 \pm 2.49$ & $[37.78, 41.34]$ \\
\hline \hline
\multicolumn{7}{|c|}{\bf{GAP VIOLATIONS}} \\
\hline
 & $\mu\pm\sigma$ & $CI_{95\%}$ & $\mu\pm\sigma$ & $CI_{95\%}$ & $\mu\pm\sigma$ & $CI_{95\%}$ \\
\hline
\textbf{FCFS} & $0.30 \pm 0.95$ & $[0, 0.98]$ & $0.20 \pm 0.63$ & $[0, 0.65]$ & $1.20 \pm 1.03$ & $[0.46, 1.94]$ \\
\hline
\textbf{IA} & $0.20 \pm 0.63$ & $[0, 0.65]$ & $0.20 \pm 0.63$ & $[0, 0.65]$ & $1.60 \pm 0.84$ & $[1.00, 2.20]$ \\
\hline
\textbf{MARL} & $0.40 \pm 0.84$ & $[0, 1.00]$ & $0.20 \pm 0.63$ & $[0, 0.65]$ & $1.00 \pm 1.05$ & $[0.25, 1.75]$ \\
\hline
\textbf{MIND-CAVs} & $0.00 \pm 0.00$ & $[0.00, 0.00]$ & $0.20 \pm 0.63$ & $[0, 0.65]$ & $0.20 \pm 0.63$ & $[0, 0.65]$ \\
\hline \hline
\multicolumn{7}{|c|}{\bf{UNNECESSARY BRAKING}} \\
\hline
\textbf{FCFS} & $1.10 \pm 0.74$ & $[0.57, 1.63]$ & $0.30 \pm 0.48$ & $[0, 0.65]$ & $1.30 \pm 1.06$ & $[0.54, 2.06]$ \\
\hline
\textbf{IA} & $1.40 \pm 0.97$ & $[0.71, 2.09]$ & $1.70 \pm 1.16$ & $[0.87, 2.53]$ & $1.60 \pm 2.07$ & $[0.12, 3.08]$ \\
\hline
\textbf{MARL} & $1.50 \pm 0.71$ & $[0.99, 2.01]$ & $1.50 \pm 1.08$ & $[0.73, 2.27]$ & $0.90 \pm 1.37$ & $[0, 1.88]$ \\
\hline
\textbf{MIND-CAVs} & $0.00 \pm 0.00$ & $[0.00, 0.00]$ & $0.40 \pm 0.70$ & $[0, 0.90]$ & $0.60 \pm 0.84$ & $[0, 1.20]$ \\
\hline
\end{tabularx}
\label{tab:combined_table}
\end{table*}

\vspace{-0.05in}
\subsection{Summary of the Results}

Across all scenarios, we observe that MIND-CAVs can (i) achieve the lowest or tied-lowest completion times; (ii) minimize the Gap violations, particularly in S1 and S3; (iii)
reduce unnecessary braking compared to IA and MARL; and (iv) have lower variance in completion time in general, indicating more stable coordination.  These results demonstrate that intent negotiation combined with centralized safety gating improves both efficiency and safety without introducing instability.

\section{Conclusion, Limitations and Future Work}
\label{sec:conclusion}

This paper introduced MIND-CAVs, a multi-intelligence negotiation framework for connected autonomous vehicles that augments conventional V2X communication with structured intent exchange and real-time edge arbitration. Unlike state-based coordination approaches that rely solely on kinematic inference, MIND-CAVs enables vehicles to explicitly communicate maneuver intent and receive safety-gated approval from a MEC node operating under bounded latency constraints.

Experimental evaluation in CARLA demonstrates that intent-driven arbitration improves both efficiency and safety in structured highway scenarios. Across multi-lane transitions and route-constrained exit maneuvers, MIND-CAVs consistently achieves lower completion times while reducing gap violations and unnecessary braking events. These results indicate that centralized intent review can stabilize multi-vehicle behavior without introducing coordination-induced latency.

%A key design objective of MIND-CAVs is real-time feasibility. Our latency analysis shows that end-to-end negotiation cycles remain below one second (335 ms for ACK decisions and 690 ms for revised PLAN responses), supporting sub-second maneuver arbitration suitable for highway driving contexts.

\vspace{0.05in}
\noindent \textit{Limitations}: The present study is limited to structured highway environments with a small number of interacting vehicles. More complex urban traffic conditions, dense multi-agent interactions, and perception noise were not evaluated. Additionally, in the current implementation, the MEC enforces a fixed 50 km/h upper speed bound during experiments. This constraint was introduced to regulate inference frequency and avoid exceeding API rate limits of the external model provider. While this does not affect comparative analysis across coordination modes, it does not reflect realistic highway speed variability.  Finally, arbitration decisions are generated using large language models operating as semantic planners rather than formally verified control policies. Although execution safety is preserved through gating, formal guarantees remain an open research direction.

\vspace{0.05in}
\noindent \textit{Future Work}: our future work will focus on four directions: (1) extending evaluation to larger fleets with heterogeneous vehicle classes and dynamic traffic density; (2) incorporating pedestrians, traffic signals, and multi-agent conflict zones; (3) removing fixed speed constraints and learning flow-aware arbitration policies that adapt to traffic conditions without inference throttling; and (4) deploying the MIND-CAV architecture on embedded edge platforms and robotic testbeds to evaluate real-world feasibility.

%By introducing structured intent negotiation and accountable arbitration into the V2X stack, MIND-CAVs provides a practical step toward transparent, real-time multi-vehicle coordination in future autonomous driving systems.

\bibliographystyle{IEEEtran}
\bibliography{references}
\end{document}